\documentclass[conference]{IEEEtran}
\IEEEoverridecommandlockouts

\usepackage{url}
\newcommand\R{{\mathbb R}}

\newcommand\N{{\mathbb N}}
\newcommand\E{{\mathbb E}}

\usepackage{cite}

\usepackage{tikz}
\newcommand{\tikzcircle}[2][red,fill=red]{\tikz[baseline=-0.5ex]\draw[#1,radius=#2] (0,0) circle ;}

\makeatletter
\newcommand\footnoteref[1]{\protected@xdef\@thefnmark{\ref{#1}}\@footnotemark}
\makeatother

\usepackage{wrapfig}
\usepackage{varwidth}
\usepackage{url}
\usepackage{graphicx}
\usepackage{booktabs}
\usepackage{amsmath}
\usepackage{amsthm}
\usepackage{amsfonts}
\usepackage{epsfig}
\usepackage{graphicx}
\usepackage{subfigure}
\usepackage{url}
\usepackage[bottom]{footmisc}
\usepackage{balance}
\usepackage{algpseudocode}
\usepackage{algorithmicx}
\usepackage{algorithm}
\algtext*{EndWhile}
\algtext*{EndIf}
\algtext*{EndFor}

\usepackage{multirow}
\usepackage{xspace}
\usepackage[singlelinecheck=off]{caption}
\DeclareCaptionType{copyrightbox}
\usepackage{pifont}
 \newtheorem{problem}{Problem}

\newcommand{\spara}[1]{\smallskip\noindent{\bf #1}}

\newcommand{\squishlist}{
 \begin{list}{$\bullet$}
  {  \setlength{\itemsep}{0pt}
     \setlength{\parsep}{3pt}
     \setlength{\topsep}{3pt}
     \setlength{\partopsep}{0pt}
     \setlength{\leftmargin}{2em}
     \setlength{\labelwidth}{1.5em}
     \setlength{\labelsep}{0.5em}
} }
\newcommand{\squishlisttight}{
 \begin{list}{$\bullet$}
  { \setlength{\itemsep}{0pt}
    \setlength{\parsep}{0pt}
    \setlength{\topsep}{0pt}
    \setlength{\partopsep}{0pt}
    \setlength{\leftmargin}{2em}
    \setlength{\labelwidth}{1.5em}
    \setlength{\labelsep}{0.5em}
} }

\newcommand{\squishdesc}{
 \begin{list}{}
  {  \setlength{\itemsep}{0pt}
     \setlength{\parsep}{3pt}
     \setlength{\topsep}{3pt}
     \setlength{\partopsep}{0pt}
     \setlength{\leftmargin}{1em}
     \setlength{\labelwidth}{1.5em}
     \setlength{\labelsep}{0.5em}
} }

\newcommand{\squishend}{
  \end{list}
}

\newcommand{\eat}[1]{}

\newcounter{ccc}

\begin{document}

\title{DCSF: Deep Convolutional Set Functions for Classification of Asynchronous Time Series}

\author{\IEEEauthorblockN{Vijaya Krishna Yalavarthi}
	\IEEEauthorblockA{\textit{University of Hildesheim}\\
		Germany \\
		yalavarthi@ismll.uni-hildesheim.de}
	\and
	\IEEEauthorblockN{Johannes Burchert}
	\IEEEauthorblockA{\textit{University of Hildesheim}\\
		Germany \\
		burchert@ismll.uni-hildesheim.de}
	\and
	\IEEEauthorblockN{Lars Schmidt-Thieme}
	\IEEEauthorblockA{
		\textit{University of Hildesheim}\\
		Germany \\
		schmidt-thieme@ismll.uni-hildesheim.de}
	
}

\maketitle

\begin{abstract}
	Asynchronous Time Series is a multivariate time series where all the channels are observed asynchronously-independently, making the time series extremely sparse when aligning them. We often observe this effect in applications with complex observation processes, such as health care, climate science, and astronomy, to name a few. Because of the asynchronous nature, they pose a significant challenge to deep learning architectures, which presume that the time series presented to them are regularly sampled, fully observed, and aligned with respect to time. This paper proposes a novel framework, that we call \underline{D}eep \underline{C}onvolutional \underline{S}et \underline{F}unctions (DCSF), which is highly scalable and memory efficient, for the asynchronous time series classification task. With the recent advancements in deep set learning architectures, we introduce a model that is invariant to the order in which time series' channels are presented to it. We explore convolutional neural networks, which are well researched for the closely related problem-classification of regularly sampled and fully observed time series, for encoding the set elements. We evaluate DCSF for AsTS classification, and online (per time point) AsTS classification.
	Our extensive experiments on multiple real world and synthetic datasets verify that the suggested model performs substantially better than a range of state-of-the-art models in terms of accuracy and run time. We increase the accuracy of the mini-Physionet dataset upto $2\%$; real datasets with synthetic setups of both AsTS, and TSMV upto 30$\%$.
\end{abstract}
\section{Introduction}
\label{sec:intro}

\begin{figure}[t]
	\centering
	\includegraphics[width=\columnwidth, clip]{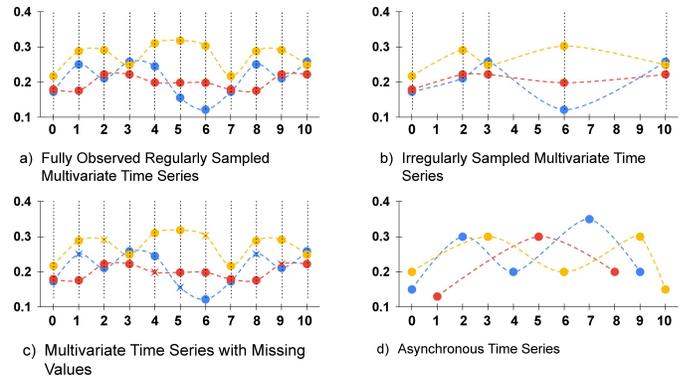}
	\caption{Demonstration of Fully observed multivariate time series (a), Irregularly Sampled Multivariate Time Series (b), Multivariate Time Series with Missing Values (c) and Asynchronous Time Series (d). observed measurements are marked with \tikzcircle[black, fill=black]{2pt} and the missing values are marked with \textbf{$\times$}.}
	\label{fig:ts}
	\vspace{-3mm}
\end{figure}

With the increase in industrialization, and new technologies, multivariate time series (MTS) datasets are becoming ubiquitous in the modern world.
Traditional deep learning models for multivariate time series classification are developed for fully observed and regularly sampled multivariate time series (RMTS) where all the variables (channels) are observed simultaneously at regular frequencies (Figure~\ref{fig:ts}(a)).
Multivariate time series where the variables are observed simultaneously but not at regular intervals are called irregularly sampled multivariate time series (IMTS) (Figure~\ref{fig:ts}(b)).
Another domain of study deals with the classification of time series that are sparse, where one or more variables are not observed at a given time point.

In some time series applications, sparsity can be introduced by missing observations, called time series with missing values (TSMV) (Figure~\ref{fig:ts}(c)). Missing values in time series are created, for example, due to sensor malfunctioning, power failures, and external physical interventions. However, in domains like medical applications~\cite{YS18}, the sparsity can be created because of asynchronous observations of the sensors which we call asynchronous time series (AsTS). The sensors are observed independent of each other, making the series extremely sparse when one tries to align them at fixed time points~\cite{YS18}. In general, AsTS have irregularly observed channels with variable lengths, and in some physiological datasets~\cite{JP16}, there might be unobserved channels as well.

In Figure~\ref{fig:ts}, we delineate the differences among all the four categories of time series. Researchers often refer to AsTS with IMTS because the samples are observed at irregular intervals. However, irregularity can happen in Univariate Time Series as well while AsTS is specific to MTS. Because of sparsity present in the time series, AsTS, and TSMV are studied under the same umbrella~\cite{CP18}. But, the qualitative difference between them might lead to performance issues in AsTSC when one uses the models that are specific to TSMV.
Though one can model AsTS as a time series with missing values, the absence of an observation in AsTS may carry its own information~\cite{LR19}, and hence, imputation schemes used for missing value time series are not always useful. This work aims at the Classification of Asynchronous Time Series AsTSC.

\begin{figure*}
	\includegraphics[width=\linewidth, clip]{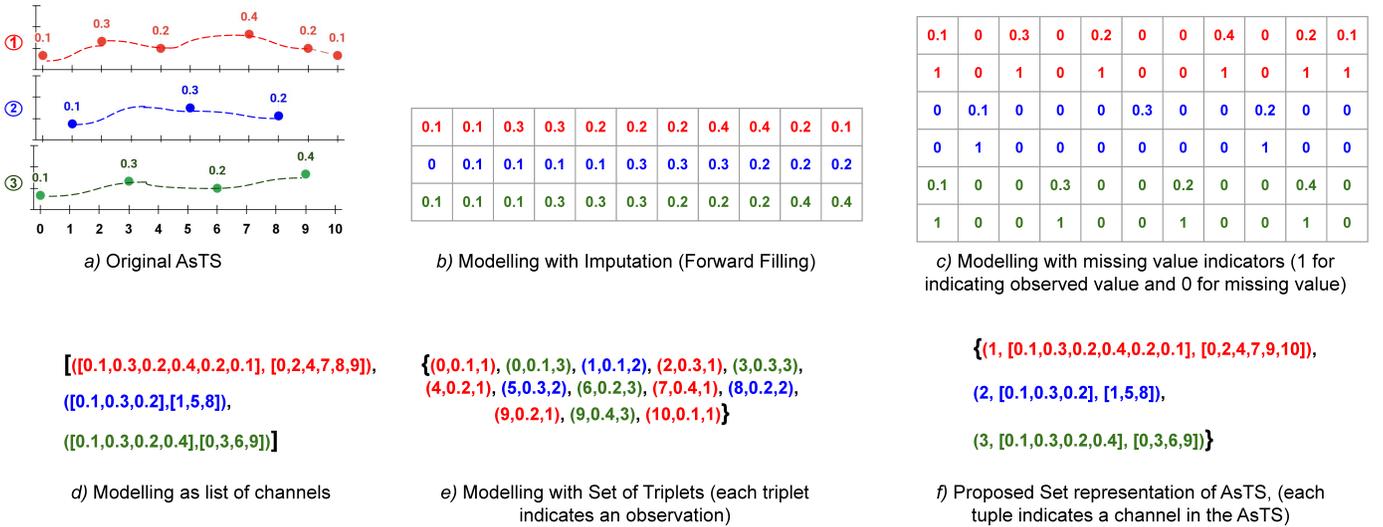}
	\caption{Demonstration of various modelings of Asynchronous Time Series.}
	\label{fig:asts_rep}
\end{figure*}

Standard Time Series Classification models assume that the time series presented to the model is RMTS, and face significant challenge in learning AsTS. In order to circumvent the problem, most of the related work model AsTS into fixed dimensions, either by aligning or using non-linear functions. They apply recurrent neural networks (RNNs) for classification~\cite{CP18,KM20,RC19,SM19,SM21}, because they can accommodate series with variable lengths. RNNs are well suited for learning the dynamics of the time series for the forecasting tasks.
On the other hand, deep learning models with convolutional layers are dominant in solving a closely related problem: classification of RMTS~\cite{RF21}. We show that by relaxing the constraint that the classification shall be performed on the synchronized channels of AsTS, using CNNs can significantly improve the accuracy of the AsTSC task.
In fact, in \cite{LC21}, Le et. al., proposed a shapelet-based model that classify RMTS by separating the channels.

Our model, Deep Convolutional Set Functions (DCSF), treats AsTS as a set of channels, and classifies it following the Deep set learning architecture~\cite{ZK17}. For this, we represent an AsTS as a set of tuples, where each tuple indicates a channel in the AsTS. In DCSF, first, we encode variable length channels into a fixed dimensional vector in a latent representation with an encoder function built using convolutional layers. Following that, we aggregate the vector representations of all the channels and classify the aggregation using a decoder. Proposed DCSF is evaluated on Physionet2012, mini-Physionet, and MIMIC datasets for the AsTS classification tasks, and used Human activity dataset for online classification of AsTS (classification per time point) task. Further, we demonstrate the performance of DCSF on synthetic AsTS datasets created from RMTS datasets.
Additionally, because one can model AsTS and TSMV in a unified manner, we verified the performance of the proposed model for TSMV using real world RMTS datasets with a synthetic setup.




The main contributions of the proposed work are:

\begin{itemize}
	
	\item We propose a novel representation of Asynchronous Time Series; set of the  Asynchronous Time Series' channels.
	
	\item We show how to apply deep set functions for time series on channels rather than individual observations.
	
	\item We perform extensive experimental analysis over multiple real and synthetic datasets; compare with state-of-the-art models for the AsTSC. We improve the accuracy of the AsTSC by $2\%$ for the mini-Physionet2012 dataset compared to state-of-the-art models with $6$ times faster run time. We improve the accuracy of the real datasets with synthetic setups of both AsTS, and TSMV upto 30$\%$.
	Experimental results attest to the superiority of the proposed model.
\end{itemize}

Our source code is provided in \\
\url{https://github.com/yalavarthivk/DCSF}
\section{Preliminaries}
\label{sec:prilim}

\begin{figure*}[ht]
	\centering
	\includegraphics[scale = 0.38, clip]{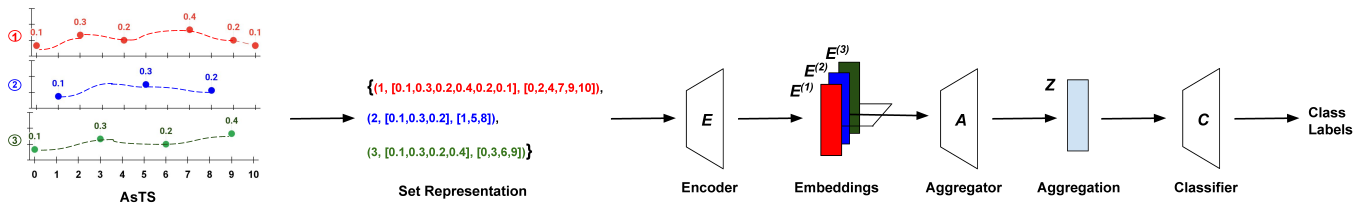}
	\caption{Proposed overall methodology}
	\label{fig:prop2}
\end{figure*}

In this section, we explain the various modeling strategies of Asynchronous Time Series (AsTS); following that, we show the proposed set representation of AsTS.

\subsection{Modeling AsTS}

In AsTS, each variable is observed independently, and the observation times are not synchronized, making the length of each channel different from others. 
Hence, it poses a significant challenge to deep learning models that operate on equal channel lengths. 
In order to train a deep learning model for AsTS data, we need to model all the channels in to equal length, or to a fixed dimensional vector.
For this, there have been various methodologies that are studied in the literature.

\spara{Data imputation:}
One can consider AsTS as a TSMV by discretizing the time into non-overlapping intervals and considering the nonexistent measurement as a missing value. Following that, these missing values are imputed using various imputation techniques. In Figure~\ref{fig:asts_rep} (a), we present an asynchronous time series where three channels are observed independently. By performing forward imputation, where the missing value is imputed with the last observed value in the channel, we achieve the fully observed time series as presented in Figure~\ref{fig:asts_rep} (b).

\spara{Missing value indicators:}
Finding a suitable imputation scheme for filling the missing values in the observation space is difficult. Hence, researchers have filled the missing observation with a constant value, generally zero, and provide a missing value indicator as a separate channel. It is assumed that the model understands the position of the missing observation through its indicator, and impute it in the latent domain. In Figure~\ref{fig:asts_rep} (c), we can see the representation of AsTS with missing value indicators. Along with the missing value indicator, some models~\cite{CP18,KM20} use time information as well for modeling AsTS.

\spara{List of channels:}
\cite{SM19,SM21} represents the AsTS as the list of channels consisting of time and observed values, as shown in Figure~\ref{fig:asts_rep} (d).  In this representation, the position of the channel is important because each channel is encoded by an independent encoder~\cite{SM21}.
It is the closest representation to the raw data of AsTS.

\spara{Set of Triplets:}
\cite{HM20} represented the AsTS as a set of observations where each observation is in a triplet form, which we call triplet notation. Each triplet carries the information of the time, value, and channel indicator. A differential set learning function is used for classifying the set. The triplet set representation is presented in Figure~\ref{fig:asts_rep} (e).



The main disadvantage of the triplets' set is that the time dependency in the series is not explicit, but has to be learned using the time information provided which is difficult, and fails when the sub-sequences carry more information of classification rather than individual observations. On the other hand, using separate encoders for every channel in list of channels approach is computationally expensive. Imputation methods are not desired when the missing observations carry the information of their own. Also, asynchronous time series come with high sparsity (our datasets have a sparsity of around 85\%), and imputing such data leads to bad predictions (we show empirically in Section~\ref{sec:exps}.
Hence, we propose a representation of AsTS (trivial to extend to any MTS) as a set of channels as shown in Figure~\ref{fig:asts_rep}(f).

\subsection{Proposed set of channels representation for Multivariate Time Series}

In contrast to the set representation in~\cite{HM20}, where
instances are represented as sets of single time slices,
we represent AsTS as set of channels, each channel being
an irregular one-dimensional time series.
Then the classification of AsTS becomes a classification
of a set of channels.

An AsTS $X$ with $D$ many channels can be represented as a set of
$D$ many elements:
  $X = \{s_1,...,s_d,..., s_D\}$.
Each of its element is a tuple $s_d = (M_d, V_d, T_d)$ of
  i) a channel indicator $M_d\in \N^P$ for channel $d\in \{1,2,...,D\}$,
     where $P$ is the dimension of the channel indicator, e.g.,
       $P=D$ for one hot encoding,
       $P=\log_2D$ for binary encoding, or
       $P=1$ for nominal encoding,
  ii) an observation vector $V_d\in \R^*$ consisting of the values observed
      in channel $d$; and
  iii) a time vector $T_d\in {\R^+}^*$ containing the corresponding
      observation times of the values in $V_d$.
We construct $T_d$ in a monotonically increasing fashion to preserve the
causal properties of the series. In our representation, the lengths of
the vectors $T_d$ and $V_d$ are always equal ($|T_d| = |V_d|$).
We denote the domain of these set elements by
  $\Omega := \N^P\times\R^*\times{\R^+}^*$ with $P\in\N$.

A time series $X$ is considered to be fully observed
if and only if
  $\forall d,e \in \{1,2,...,D\}, T_d = T_e$,
meaning all the variables are observed at any given time point.
Similarly, a time series $X$ is considered to be sparse,
if there exists at least two variables $d$ and $e$
such that $T_d\ne T_e$.
Note that our representation covers both, TSMV and AsTS,
although they are qualitatively different.
In Figure~\ref{fig:asts_rep} (f), we demonstrate our set
representation for the AsTS of Figure~\ref{fig:asts_rep} (a).

\newcommand\D{{\mathcal D}}
\newcommand\train{^{\text{train}}}
\begin{problem}[Asynchronous Time Series Classification]
Given i) a data set $\D\train$ of elements 
           $(X,Y)\sim \rho$
             sampled from an unknown distribution $\rho$ on $\Omega^*\times\{0,1\}^L$
            ($L\in\N$), and
      ii) a loss function $\mathcal{L}: \{0,1\}^L\times\R^L \rightarrow \R$,
            e.g., cross entropy,
find a model $f: \Omega^* \rightarrow \R^L$
with minimal expected loss:
     $\min_{(X,Y)\sim \rho} E(\mathcal{L}(Y, f(X)))$.
\end{problem}

\section{Proposed Model}

Motivated by the deep set learning functions~\cite{ZK17}, we propose DCSF for the classification of AsTS. In the following, we explain the model architecture, and its building blocks.



\subsection{Model Architecture}
\label{sec:mod_arch}

We define $F$ as a function that operates on set elements, and its output is invariant to the order in which the set elements are presented. The proposed model has three different modalities: 1) Encoder ($E$), 2) Aggregator ($A$) and 3) Classifier ($C$). We formulate $F$ for an AsTS instance $X$ with $D$ many set elements (channels) as follows:

\begin{align}
F(X) = C(A(E(s_1), ..., E(s_D))), \quad \quad s_d \in X \label{eq:deepset}
\end{align}

where $E: \Omega \rightarrow \R^{K}$, $A: \R^{K\times D} \rightarrow \R^{K}$and $C: \R^K \rightarrow \R^L$ where $K \in N^+$ is the dimensionality of the latent representation of a set element $s_d \in X$. The overall architecture of the proposed model is presented in Figure~\ref{fig:prop2}.

\spara{Encoder ($E$)}

\begin{figure}[ht]
	\centering
	\includegraphics[width = 0.9\columnwidth, clip]{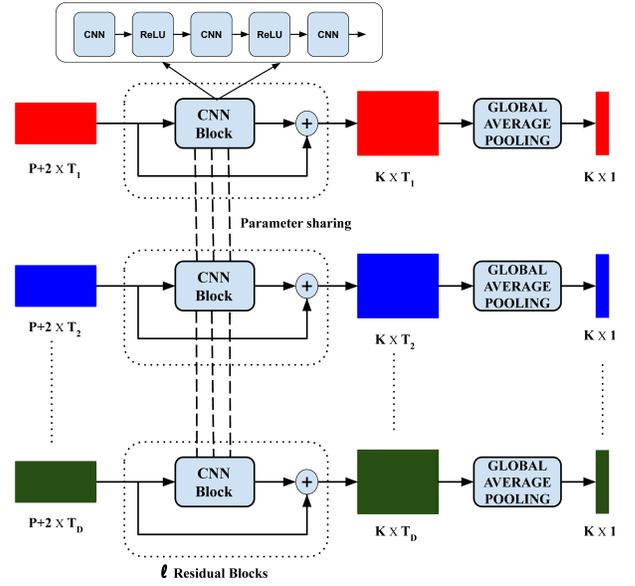}
	\caption{Encoder (\textit{E}) architecture. $l$ residual blocks are used for extracting the latent embeddings from a set element (a channel of AsTS). Parameters are shared across all the element. We aggregate the variable length latent embedding into a fixed dimension using Global Average Pooling.}
	\label{fig:prop1}
\end{figure}

Our encoder $E$ takes a set element as input, and provides a fixed dimensional representation $\R^K$.
Since the length of every channel varies in AsTS, we require a model whose parameters do not depend on the length of the series presented to it.
The widely used parameterized model architectures for this task are Recurrent Neural Networks (RNN), Attention Network (AtN), and Convolutional Neural Networks (CNNs).
While RNN and AtN are useful for time series forecasting, state-of-the-art models for time series classification are built upon CNNs\cite{RF21}.
Hence, we choose a CNN based model as an encoder, and extract useful information from the channel observations.
The proposed encoder architecture is presented in Figure~\ref{fig:prop1}. It comprises a series of ${l}$ many residual blocks where each residual block consists of three convolutional layers with a ReLU activation function following each layer. In order to achieve fixed representation, one needs to aggregate the latent representation over time. For this, we employ global average pooling, which is widely used in CNNs based time series classification~\cite{RF21} models.

As mentioned in Section~\ref{sec:prilim}, our set elements are tuples. Since our encoder cannot directly operate on tuples, we convert our set element $s_d$ into a multi-dimensional array $\mathtt{s}_d \in \R^{(P+2)\times |T_d|}$. $\mathtt{s}_d^{:,t} = [M_d, V_d^t, T_d^t], t\in\{1,...,|T_d|\}$ consists of 1) modality indicator of channel $d$, ($M_d\in N^P$), 2) observed value at $t$ ($V_d^t$), and 3) its observation time ($T_d^t$).
Our encoder $E$ takes $\mathtt{s}_d$ as an input, and outputs $K$ dimensional fixed representation of $\mathtt{s}_d$. We consider $K$ to be large enough in order to memorize the temporal locations of the signals present in $s_d$. In~\cite{WF19}, Wagstaff et. al., shows that  universal function representation is guaranteed when $K\ge \max_{i} |X_i|$ meaning, the dimension of latent embedding shall be greater than or equal to the number of set elements in $X$. We carefully choose $K$ that follows the criteria using hyperparameter search.

The proposed encoder share some similarity with the ResNET architecture provided in~\cite{RF21} in terms of using residual blocks, number of kernels, and kernel lengths in CNNs used in the residual block. However, a) we do not have a fixed number of residual blocks meaning, the number of residual blocks ${l}$ is a hyper parameter, b) we do not use batch normalization after each residual block as it is hindering the performance of DCSF (refer Section~\ref{sec:ablation}).

%
%
\spara{Embeddings Aggregator ($A$)}

Once the embeddings for each $s_d\in X$ are computed, we aggregate those embeddings $E(s_d)$ using the aggregator function $A$. There have been various kinds of aggregations provided in the literature such as mean aggregation~\cite{ZK17,GR18}, sum aggregation~\cite{ZK17,GR18}, and attention based aggregation~\cite{HM20}. In this work, we sum all the embeddings in order to aggregate them as shown in Equation~\ref{eq:agg_sum} and represent the aggregated value with $z$.

\begin{align}
\label{eq:agg_sum} 
z = A\left(E\left(s_1\right), ..., E\left(s_D\right)\right) = \sum_{d=1}^{D}E(s_d), \quad s_d \in X
\end{align}

%

\spara{Classifier ($C$)}

After aggregation of the embeddings, we receive a vector representation $z \in \R^K$ of the multivariate time series. Then, one can use any model that is used for vector data classification. In this work, we use a fully connected artificial neural network.

%



As mentioned earlier, SEFT-ATTN presented in~\cite{HM20} also uses a deep set architecture for classifying time series. However, SEFT-ATTN uses triplet set modeling of time series, and those triplets are given as inputs to the encoder; an attention mechanism was used for aggregating the latent embeddings of the set elements. We observed that operating on individual observations has the disadvantage of losing sequence information even after providing time in the triplet. On the other hand, operating on sequences rather than observations not only preserves the causality of the sequence but also eases the learning process.



\subsection{Online classification scenario}
\label{sec:online_class}

For the online classification we need to predict the class label at every time point. Hence, we replace normal convolutions with causal convolutions, such that the current observation cannot look ahead into the future. Because we need to classify at every time point, we cannot implement global average pooling as it is implemented on entire sequence. Instead, we use causal average pooling ($CAP$) where at every time point we average the observations made until then, and aggregate those embeddings followed by classification using dense layer. The causal average pooling at time point $\tau$ for an array $\mathcal{E}$ is given by
\begin{equation}
CAP(\mathcal{E}^\mathcal{\tau}) = \frac{1}{\tau}\sum_{i=1}^{\mathcal{\tau}} \mathcal{E}^i
\end{equation}

\subsection{Supervised learning}

We consider the parameters of the encoder $E$ and the classifier $C$ as $\theta$ and $\phi$, respectively. Because we perform sum aggregation, we do not have any parameters for $A$.
We follow~\cite{ZK17}, and train the model with all the channels present in time series. We optimize the following cost function:

\begin{equation}
\mathcal{J} := \E_{(X,Y)}\left[\mathcal{L}\left(Y, C\left(A\left(E\left(X;\theta\right)\right);\phi\right)\right)\right]
\end{equation}

where, $\mathcal{L}$ is the loss function which is binary cross entropy for binary classification and Sigmoid cross entropy for multi-class classification tasks.

\section{Experiments}
\label{sec:exps}


We implemented DCSF using TensorFlow and ran all the models on Nvidia GeForce RTX 3090 GPU nodes. In order to promote reproducibility, we outsource our source code in \\ \url{https://anonymous.4open.science/r/AsyncTSC-3561}.
 

\subsection{Asynchronous time series Datasets}
\label{sec:meddata}

Following the literature that deals with AsTSC problems, we chose three medical and one Human Activity Recognition dataset that contain asynchronous measurements. Basic statistics of the datasets are presented in Table~\ref{tab:data_stat}.

\spara{Physionet:}

The PhysioNet Challenge 2012 dataset~\cite{SM12,GA00} comprises asynchronous time series data extracted from the records of patients admitted to ICU. Up to 37 variables were measured for the first 48 hours after the patient's admission, and all the time series have general descriptors like height, weight, age, and gender. The aim of this dataset is to predict whether a person dies in the hospital. The dataset comprises three different splits (a, b, and c) with 4000 observations each. Here, we combine all the splits, round the observed time points to the nearest minute, and split the 12,000 labeled observations randomly into 80\% for training and 20\% for testing.

\spara{mini-Physionet:}

Baseline models~\cite{SM21,RC19} used only a part of the full dataset for the experiments. Hence, we would like to see how the model performs for a small part of the dataset. For this, we have taken split-c of the physionet dataset. It is also an imbalanced dataset with 4000 samples. Again, we randomly split $80\%$ of the dataset into training, and $20\%$ for testing.

\spara{MIMIC:}

The MIMIC dataset~\cite{JP16} also comprises records of ICU patients at Beth Israel Deaconess Medical Center. All the time series contain asynchronous measurements. Again, the task is to predict patient's mortality after the first 48 hours of ICU admission. We follow the procedures of~\cite{HM20,HK19} for splitting the dataset that contains around 21,000 stays with 12 physiological variables being observed.

\spara{Activity:}

We use the human activity dataset for the online classification of Asynchronous Time Series. It contains the AsTS collected from five individuals wearing 3D sensors positioned in the belt, chest, and ankles while performing various tasks like sitting, laying, walking, etc. The dataset comprises $6,554$ time series with $12$ channels, and channel consists of $50$ time points.
We follow the data preprocessing steps mentioned in~\cite{SM21,RC19}. The task is to classify every time point in the series into $11$ classes (activities).

\begin{table*}[ht]
	\centering
	\small
	
	\caption{Comparison of the results over the medical datasets namely Physionet2012, mini-Physionet2012, and Mortality (MIMIC) in terms of AUROC and run time for AsTSC task. Best results are presented in bold, and the second best are presented in italics.}
	\begin{tabular}{|l|c|c|c|c|c|c|}
		
		\hline
		\multirow{2}{*}{Model}  & \multicolumn{2}{|c|}{Physionet} & \multicolumn{2}{|c|}{mini-Physionet} & \multicolumn{2}{|c|}{MIMIC}\\
		\cline{2-7}
		& \multicolumn{1}{c|}{AUROC}                  & Time
		& \multicolumn{1}{c|}{AUROC}                  & Time & \multicolumn{1}{c|}{AUROC} & 
		{Time}  \\
		\hline
		GRU-D               &86.3$\pm$0.3   &       104s                & 83.2$\pm$0.3       & 24s                      &  \textit{85.6$\pm$0.2}     & 430s                     \\
		IP-NETS               &86.5$\pm$0.2&    14s                     & \textit{84.2$\pm$0.4} & 6s                       & 84.2$\pm$0.5         & 297s           \\
		Phased-LSTM             &     79.6$\pm$0.4 &       70s            & 78.9$\pm$0.3        & 46s         &   78.9$\pm$0.4       & 452s                      \\
		Latent-ODE                &85.7$\pm$0.7$^*$&  3500$^*$               &        82.9$\pm$0.4$^*$              &     1200s                                   &  80.9$\pm$0.2$^*$                &             4600$^*$  \\
		SEFT-ATTN                   &86.0$\pm$0.4&   7s            &   81.8$\pm$0.4    & 4s                       &  85.0$\pm$0.2         & 29s           	    \\
		
		mTAN                          &\textit{86.7$\pm$0.0}&     6s       & 84.0$\pm$0.4  &                     {{5s}}     & 83.3$\pm$0.1         &     55s            \\
		\hline
		ResNET                          &84.2$\pm$0.7&    8s       & 77.8$\pm$0.7 &        6s                       & 81.5$\pm$0.6            & 		25s	\\
		ResNET-forw.&82.6$\pm$0.9&10s&77.4$\pm$2.2&19s&82.3$\pm$1.4&51s\\
		\hline
		DCSF (Ours)     &\textbf{87.1$\pm$0.2}&              14s             & \textbf{86.2$\pm$0.3}& 1s                      & \textbf{85.8$\pm$0.1}  &   11s               \\ 
		\hline
	\end{tabular}
	\label{tab:med_data_results}
\end{table*}

\subsection{Competing models}

We use the following baseline models for the comparison.

\spara{GRU-Decay (GRUD)}~\cite{CP18} proposed modifications to the hidden state of the GRU cell allowing a learnable decay rate to decay the past observations to the mean of the variable.

\spara{Phased-LSTM}~\cite{NP16} originally proposed for irregularly sampled time series, but not for unaligned measurements. We perform forward filling in order to handle the partially observed time points.

\spara{Interpolation Networks (IP-NETS)}~\cite{SM19} use several interpolation
layers, followed by a GRU. The interpolation layers are learned along with a classifier in an end to end fashion.

\spara{LatentODE}~\cite{RC19} proposed model with the encoder as an ODE-RNN and decoder as a neural ODE.

\spara{Seft}~\cite{HM20} is a deep set model that uses triplet set modeling of time series. It extracts embeddings from all the observations, performs weighted average using an attention mechanism, and applies a fully connected network for classification.

\spara{mTAN}~\cite{SM21} proposed an attention-based interpolation for modeling AsTS into a fixed dimension. Later, the Variational AutoEncoder based encoder-decoder module was used for Classification (main task) and Reconstruction (auxiliary task) of AsTS.

\spara{ResNET} is a Time series Classification model that uses CNNs presented in~\cite{RF21}. We fill the missing values with zeros and provide missing value indicators as additional channels in the series.


\spara{ResNET-forward} is also a ResNET model~\cite{RF21}, we fill the unobserved value with the last observed one in the respective channel (forward filling).

\spara{DCSF} is our proposed model. In DCSF we use one hot encoding of variable as channel indicator. We normalized the observation times using min-max scaling.

\subsection{Experimental Protocol}
\label{sec:exp_prot}

We randomly split off $20\%$ of training data for validation, which is used for hyperparameter search and early stopping. 
For training, we use Adam optimizer with learning rate chosen from $[0.001, 0.00001]$ and the batch size from $\{32, 64, 128\}$. We also consider normalizing the time series as a hyperparameter because, for some datasets, normalization of the series before inputting to the models provides better validation results. We set the embedding length $K=128$ in order to follow the universal function approximation for set functions presented in~\cite{WF19}.

The hyper parameters searched for all the competing models are provided in the supplementary material.

We randomly generate $10$ sets of hyperparameters for each of the competing models, and chose the setup with the best evaluation metric on the validation dataset. The models with selected hyperparameters are run independently for $5$ times. For the medical datasets, we use area under receiver operating characteristic (AUROC) as the evaluation metric, because those datasets are heavily imbalanced. For the Activity dataset, we use accuracy as the evaluation metric.

\subsection{Experiments on Asynchronous Time Series Classification}

We present experimental results for classification of full AsTS using the medical datasets in Table~\ref{tab:med_data_results}. The presented run time is the one that is taken by the model with best chosen hyperparameters. It could be possible that different hyperparameters may need different run times, but we use the current set up in order to be fair for all the models. 

We can observe that our proposed model outperforms all the baselines in the Physionet and mini-Physionet datasets. Whereas, in the MIMIC dataset, our proposed model performs on par with the GRU-D model while outperforming all the remaining baselines. However, the run times of our proposed model are less than the GRU-D by an order of magnitude for the MIMIC dataset.
Note that the second-best performing model is the Interpolation Networks, and the proposed model improves the AUROC of Interpolation Networks by around $2\%$ in the mini-Physionet and MIMIC datasets. The Latent-ODE model could be slightly underperforming, because we run the dataset with the default hyperparameters provided in~\cite{RC19}. This is because of high computational complexity and run times. Especially, for the MIMIC and Physionet datasets, it took around an hour for each epoch, which is also observed in~\cite{HM20}.

The results of the Physionet and MIMIC dataset are similar to that of the results published in~\cite{HM20}, whereas the results on mini-Physionet dataset deviate significantly from the published results in~\cite{SM21}. Our results, in most cases, are better than the published ones because, unlike the procedure in~\cite{SM21,RC19}, we balance both the positive and negative class time series while sampling a batch for training in order to avoid bias towards the negative samples. Also, we could not reproduce the results published in~\cite{SM21} with the provided experimental setup and hyperparameters. We emailed the authors regarding the issue, and did not receive any information from them. The same issue has been raised by multiple people regarding their inability to reproduce the results for both Physionet and MIMIC datasets.~\footnote{\url{https://github.com/reml-lab/mTAN/issues}}.

\subsection{Experiments on Online Classification}
\label{sec:exp_act}

The Activity dataset is an online classification dataset where one needs to provide class labels at every time point.
While this is similar to segmentation in the time domain, since many researchers have explored this task, we want to see how the proposed model works in this scenario. From Table~\ref{tab:activity}, it can be observed that the proposed model outperforms all the baselines significantly, and the only model that performs close to ours is mTAN~\cite{SM21}. Since the source codes are not available for multiple baseline models, we did not run the experiments for the published baseline models, but took the results from~\cite{SM21}. Also, we could not compare the model in terms of run time because, the information is not available in the published works~\cite{SM21}.

 \begin{wraptable}{l}{4cm}
	\small
	\caption{Results for the Activity dataset}
	\label{tab:activity}
	\begin{tabular}{|l|c|}
		\hline
		Model & Accuracy \\
		\hline
		GRU-D &  86.2$\pm$0.5\\
		IP-NETS    & 86.9$\pm$0.7\\
		Phased-LSTM    &  85.5$\pm$0.5\\
		Latent-ODE     &   87.0$\pm$2.8\\
		SEFT-ATTN      &  81.5$\pm$0.2 \\
		mTAN       &  \textit{91.0$\pm$0.2} \\
		\hline
		ResNET &  88.7$\pm$0.2 \\
		ResNET-forw. & 87.1$\pm$0.4  \\
		\hline
		DCSF (Ours) &  \textbf{91.3$\pm$0.1} \\
		\hline
	\end{tabular}
	\vspace{-3mm}
\end{wraptable}


One important observation is that the Activity dataset does not heavily depend on the previous observations while predicting the class label at the current time point. CNN-based models under perform when the kernel lengths are larger than $1$.
Hence, we set the kernel length of CNNs to $1$ for the CNN-based models, namely RESNET, RESNET-forward, and our DCSF.

\subsection{Experiments on Regularly sampled Multivariate Time Series (RMTS) Datasets with asynchronous setup}
\label{sec:synthdata}

Since the medical datasets were used by all the baseline models in their works, it is possible to have a model bias towards those datasets. Hence, we additionally use two RMTS datasets with synthetic AsTS setup for the comparison.

We consider two RMTS datasets, namely LSST and Phoneme Spectra~\cite{RF21}, and induce artificial sparsity to them. These are among the $4$ largest datasets used in~\cite{RF21}. The largest dataset Pen Digits, has very few time stamps (8) and channels (2) in a series. In the second largest dataset Face Detection, the accuracy of the best classifier in asynchronous setup is close to the default rate ($50\%$) of the dataset. Hence, we choose third and fourth largest datasets.

\begin{table}[ht]
	\centering
	\small
	\caption{Comparison of the results over asynchronous setup on RMTS datasets: Phoneme Spectra and LSST. Sampled a single variable for every time point. Clas. full data is the classification result of the ResNET model~\cite{RF21} when the full (RMTS) dataset is used.}
	\label{tab:synthsa}

	\begin{tabular}{|l|c|c|}
		\hline
		
		Model & {Phoneme Spectra} & {LSST}\\
		\hline
		GRU-D                                      &	08.2$\pm$0.4	&	41.8$\pm$0.8\\
		IP-NETS                                    &	\textit{11.6$\pm$0.5}		&  44.8$\pm$0.7     \\
		Phased-LSTM	&	04.3$\pm$1.2	&	40.7$\pm$0.6	\\
		Latent-ODE                                 &           09.3$\pm$0.7            &    38.5$\pm$0.5\\
		SEFT-ATTN                                  &	10.4$\pm$0.4	&  \textit{46.6$\pm$0.6}              \\
		mTAN                                       &	09.1$\pm$0.1	&	40.7$\pm$0.8	\\
		\hline
		ResNET                                     &        10.2$\pm$1.2   & \textbf{47.8$\pm$0.6}\\
		ResNET-forward &	10.1$\pm$1.0	&47.3$\pm$1.2\\
		\hline
		DCSF        (Ours)                       & \textbf{13.4$\pm$0.3} &      {46.1$\pm$0.5}          \\
		\hline
		Default rate    & 2.6   & 31.5  \\
		Class. full data & 31.8 & 70.19 \\
		\hline
	\end{tabular}
\end{table}

We use the following setup to generate asynchronous datasets from RMTS. Because in an asynchronous dataset variables are observed independently, we assume that at every time stamp only one variable is observed. Hence, in the synthetic setup we choose one variable uniformly at random for a given time point. The total number of observations is equal to the length of the RMTS. For any two different samples, the observation times of a channel may not be consistent.

The experimental results on the datasets with synthetic asynchronous setup are presented in Table~\ref{tab:synthsa}. Accuracy is the evaluation metric. For Phoneme Spectra, proposed model outperforms all the baseline models by a significant margin; the accuracy improvement is around $15\%$ compared to the next best model, Interpolation Networks. For the LSST data, imputation based models, the ResNET-forward, and the RESNET perform slightly better than the DCSF. SEFT-ATTN has an accuracy gain over DCSF, but the difference is not statistically significant.


\subsection{Ablation study}
\label{sec:ablation}

\begin{table}
	\centering
	\small
	\caption{Comparison of the DCSF model with i) model that uses best channel for classification ii) ensemble model of all the channel classifiers. Evaluation metric is AUROC.}
	\label{tab:exp_sing_ens}
	\begin{tabular}{l|c|c|c}
		Dataset & DCSF & Single Channel & Ensemble \\
		\hline
		\hline
		Physionet   &   87.1$\pm$0.2&   83.0$\pm$0.1    &   76.4$\pm$0.0\\
		mini-Physionet &	86.2$\pm$0.3	&	83.2$\pm$0.2	&	71.0$\pm$0.2\\
		Mimic	&	85.6$\pm$0.1	&	71.3$\pm$1.1	&	78.2$\pm$0.6\\
		LSST	&	48.2$\pm$0.7	&	41.7$\pm$0.7	&	42.9$\pm$0.6\\
		Phoneme Spectra	&	13.4$\pm$0.4	&	7.6$\pm$0.4	&	9.7$\pm$0.5	\\
	\end{tabular}
\end{table}

\spara{Comparison with models classifying single channel:}

Since we separate the channels before inputting to the encoder, one might assume that the model is not learning from channel interactions, but just yields the accuracy of a channel that contributes maximum to the output. Hence, we perform an experiment by providing only a single channel to the model, and compared it with the proposed model. We ran the experiments with all the channels, and present the results for the best one in Table~\ref{tab:exp_sing_ens}. From the experimental results, we can conclude that the proposed model does learn from the cross channel interactions as it heavily outperforms the model that learns from only a single channel. One important observation is: for the mini-Physionet dataset, learning a single channel provides directly better results compared to many state-of-the-art models.

\spara{Comparison with an ensemble model:}

In order to demonstrate that the aggregation after computing the latent embeddings is useful in learning the model, we compare the proposed model with an ensemble model. For this, we learn $D$ models for $D$ many channels present in the AsTS data. Each model takes the series from a single variable, and is trained for the classification. While testing, we take the average of encodings of the penultimate layer (while Softmax is the last layer), and compute the class label.

We present the results in Table~\ref{tab:exp_sing_ens}. We observe that the ensemble model provides better results than the model that learns with a single channel, except for the Physionet datasets. The reason for this phenomena is, the channels other than the best one do not provide useful result on their own. Hence, the linear combination used for ensemble, suppress the result of the best channel.
Moreover, the proposed model outperforms the ensemble model by a significant margin, showing the advantage of aggregating the embeddings after the encoder.

\begin{figure}[t]
	\centering
	\includegraphics[width=\columnwidth, clip]{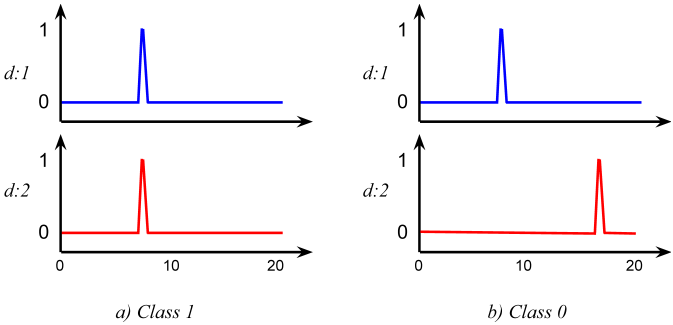}
	\caption{Toy data used for the ablation study to find the importance of time information in the encoder input. Positive class is shown in a) where peaks are observed at same time. Negative class is shown in b) where peaks are observed at different time points}
	\label{fig:toydata}
\end{figure}

\spara{Experiment to verify the importance of time information:}

The purpose of having time information as a dimension in the encoder input is to capture the time dependent channel interactions. Because we are encoding the channels separately, one might doubt that the proposed model is not capturing the channel interactions. In order to show that the time information provided in the encoder input serves the purpose, we performed an ablation study with a toy dataset, where the time information, or channel interactions over time is absolutely necessary for classifying it.

In the toy dataset, each time series has two channels of length $T$, and they are initialized with $0$ for all the time steps. For positive samples, we sample a time location $t\in \{0,1,...,T\}$, and substitute $1$ for both the channels at time point $t$. Where as, for negative samples, we sample two locations $t_1, t_2 \in \{0,1,...,T\} (t_1 \ne t_2)$ and replaced $0$ with $1$ at $t_1$ for channel $1$, and $t_2$ for channel $2$ as shown in Figure~\ref{fig:toydata}. For this experiment, we set $T=20$, and induced sparsity of $10\%$, $50\%$, and $90\%$ without altering the signals.

We conducted experiments on the proposed model with and without time information, and the results are presented in Table~\ref{tab:toy}. We can observe that the proposed model performs better with time information (DCSF) compared to the one without (DCSF w$\setminus$o time). It shows that the proposed model can learn the channel correlation towards the output when explicit time information is provided to it.
An interesting observation is, the proposed model without time can learn from the relative position of the signal when the data is less sparse. With an increase in sparsity, the performance reduces. We can see that for $10\%$ and $50\%$ sparsity, the model achieves $91\%$ and $76\%$ accuracy, respectively.

\begin{table}[t]
	\centering
	\small	
	\caption{Comparison of the results on toy dataset for the proposed model with and without time information.}
	\begin{tabular}{|l|c|c|c|}
		\hline
		& 10\%              & 50\% & 90\%\\
		\hline
		DCSF w$\setminus$o time                                 & {91.1$\pm$0.9} & 76.4$\pm$0.8                       & 59.5$\pm$0.6\\
		DCSF (Ours) & \textbf{99.1$\pm$0.2} & \textbf{97.0$\pm$0.6}                       & \textbf{97.6$\pm$0.7}\\
		\hline
	\end{tabular}
	\label{tab:toy}
\end{table}

\spara{Experiments using various time embeddings:}

Here, we study multiple time embeddings in order to embed time information into the model.

\textit{Absolute time information (TE-1)}
Here, we provide the normalized absolute time values directly to the model, and the embeddings are learned in the latent layers, which is the proposed modeling.

\textit{Time differences (TE-2)}
We consider the distance between the current observation and the previous one in time instead of absolute time information.

\textit{Sinusoidal positional embedding (TE-3)}
Here, we implement sinusoidal positional embedding provided in~\cite{HM20}, which is a variant of the one presented in~\cite{VS17}.

\textit{No Time embedding (TE-4)}
It is interesting to see how the models work when there is no information about time is given to it. Hence, we did not provide any explicit time information to the model.

\begin{table}
    \small
    \caption{Ablation study with various model configurations}
    \label{exp:model_config}
    \centering
    \begin{tabular}{l|c|c}
    Configuration   &   mini-Physionet   & MIMIC \\
    \hline
    TE-1 (DCSF)    &  86.88 & 85.60  \\
    \hline
    \hline
    TE-2        &   86.35   &   85.40   \\
    TE-3        &   82.59   &   82.04   \\
    TE-4        &   86.31   &   85.12   \\
    \hline
    \hline
    DCSF+BN      &   84.69   &     85.15  \\
    DCSF+IE      &   86.35   &   85.42    \\
    \end{tabular}
\end{table}

Results on the validation dataset are provided in Table~\ref{exp:model_config}, shows that the absolute time values are more helpful compared to all the other embeddings. Another observation is that, model provides good accuracy even with no explicit information. This makes us think that the datasets we are using may not have significant time dependency. However, when there is a time dependency, our model excels as shown in the previous section.

\spara{Experiments with different model configurations:}

Here, we study multiple model configurations and the results for the MIMIC and mini-Physionet are presented in Table~\ref{exp:model_config}. It can be seen that the batch normalization (DCSF+BN) is not helpful in the proposed model. We compare DCSF with a model where each channel is trained with an independent encoder (DCSF+DE). When we use $D$ many encoders for $D$ many channels, the channel interactions entirely depend on the aggregated embeddings. In the DCSF, because of the shared parameters, the encoder also receive the channel interactions, and with very few parameters.

\begin{table*}[ht]
	\centering
	\small
	
	\caption{Comparison of the results over two RMTS datasets namely Phoneme Spectra and LSST with missing value setup. Artificially induced $p\%$ sparsity to the dataset by randomly removing $p\%$ of the observations. We set $p = 10, 50, 90$.}
	\label{tab:missing}
		\begin{tabular}{|l|c|c|c|c|c|c|}
		\hline
		
		\multirow{2}{*}{Model}& \multicolumn{3}{c|}{Phoneme Spectra} & \multicolumn{3}{c|}{LSST}\\
		\cline{2-4} \cline{5-7}
		& 10\%              & 50\% & 90\% & 10\%                  & 50\% & 90\% \\
		\hline
		GRU-D                              &        20.8$\pm$1.2	& 15.8$\pm$0.6          &     8.6$\pm$1.7      &	59.4$\pm$1.0	& 51.2$\pm$0.5              &        41.1$\pm$0.8               \\
		IP-NETS                              &	21.1$\pm$0.6	& 16.7$\pm$0.6          &	10.4$\pm$0.7	& 61.8$\pm$0.2 & {59.0$\pm$0.3}     & 41.0$\pm$1.2                      \\
		Phased-LSTM                           &	17.1$\pm$0.6	&	06.2$\pm$0.3          &	4.5$\pm$0.4  &	57.8$\pm$0.6	& 48.8$\pm$1.1              &    40.3$\pm$0.7\\
		Latent-ODE                             &  12.3$\pm$1.3  &   09.2$\pm$0.8&   08.2$\pm$0.4                         & 38.3$\pm$2.0 &  36.6$\pm$1.6  &   39.3$\pm$0.7\\
		SEFT-ATTN                               &	10.5$\pm$0.5	& 10.9$\pm$0.7          &	11.0$\pm$0.5		&60.0$\pm$2.1	&	47.3$\pm$1.3	&	\textbf{46.7$\pm$0.7}	\\
		mTAN                                     &	11.9$\pm$0.6	& 11.8$\pm$0.5          &     7.4$\pm$0.3              &	39.9$\pm$1.7	&	39.1$\pm$0.4	&	35.7$\pm$0.9\\
		\hline
		ResNET                                 &    22.5$\pm$0.2    & 12.7$\pm$0.6          &   7.1$\pm$1.4 &   \textit{65$.0\pm$1.4}& {{58.7$\pm$2.2}}     &   44.1$\pm$1.7\\
		ResNET-forw.	&	\textit{29.0$\pm$0.7}	&	\textit{19.5$\pm$0.7}	&	\textit{11.4$\pm$0.2}	&	64.5$\pm$2.2	&	\textit{59.1$\pm$1.3}	&	\textit{44.7$\pm$0.4}\\
		\hline
		DCSF (Ours)	& \textbf{31.5$\pm$1.1}	&	\textbf{25.8$\pm$0.7}	& \textbf{14.4$\pm$0.8}	&	\textbf{65.8$\pm$1.4}	&	\textbf{{60.2$\pm$0.4}}	&	{43.8$\pm$0.9}\\  
		\hline
	\end{tabular}
\end{table*}

\subsection{Experiments on time series with missing values dataset}
\label{sec:exp_synth_mcr}

As mentioned earlier, both time series with missing values and asynchronous time series can be modelled in a unified manner. Hence, it would be interesting to see the performance of the competing models in the missing value setup. For this, we remove $p\%$ of the observations from the RMTS. As an example, if an RMTS has $5$ variables observed for a length of $50$, there will be $250$ observations. For every time series, we randomly remove $p\%$ of those $250$ observations in order to create TSMV.

We present the results on the LSST, and Phoneme Spectra with $p \in \{10, 50, 90\}$ in Table~\ref{tab:missing}. With accuracy being the evaluation metric, the best results are presented in bold and the second-best in italics. We observed that the proposed model performs better among all the competing models. The next best model is ResNET-forward which is an imputation model. For Phoneme Spectra, with $p=50\%$, DCSF provides $29\%$ better accuracy compared to baseline models. Results show that, compared to the medical datasets, the lifts in the synthetic setup are significant, because for the medical datasets, the required information for classification is provided, but the observations are not synchronized. Whereas for the synthetic setup, we randomly remove the values, making it difficult for the models to learn. {\em The results indicate that the proposed DCSF can be used for both AsTS and TSMV datasets}.

Again, for the setting of $90\%$ sparsity for LSST dataset, DCSF performs worse than ResNET, ResNET-forward and SEFT-ATTN as seen in the synthetic asynchronous setup for the LSST dataset. We observe that with extreme sparsity, in the LSST dataset, the sequence information is destroyed and applying convolutions on the observations with large time gap is not useful.

\section{Literature Review}
\label{sec:related}

This work focus on the problem of classification for asynchronous time series (AsTS) data where the variables of the series are observed independent of each other. We briefly discuss the recent works that studied this problem.


Though one can consider the classification of both time series with missing values (TSMV), and AsTS as closely related problems, there is a qualitative difference between them. Missing values are observed due to the malfunctioning of a sensor, whereas in AsTS all the sensors work independently.
The time axis can be discretized into non-overlapping intervals, and consider the intervals with no observations as missing values~\cite{BL19} making AsTS extremely sparse. Data imputation schemes or missing value indicators are used for the classification. For example, \cite{MK12} performed semi-supervised clustering of medical data using Gaussian mixture models. Later, \cite{LK16} discretized the time axis into hour-long bins, aggregated the information, and passed it through an RNN along with the missing value indicators. Chen et. al.~\cite{CP18}, proposed various methods by combining Gated Recurrent Units (GRUs), and imputation schemes along with the one that takes observed values, missing value indicators, and the time difference between two observations. Especially in GRU-Decay, the last observed value is decayed to the mean value that is learned while training. Even though these approaches can be implemented for AsTS classification, they depend on the imputation of the time series data in input or latent domains rather than directly using the data for classification.


Rather than modeling AsTS as a TSMV, researchers developed models that can directly work on AsTS. In~\cite{CR18}, Chen et. al., propose a Variational Auto-encoder based model that uses a neural network based decoder model combined with a latent ordinary differential equation model, for the continuous time series. Time series data is modeled using a continuous time function in the latent domain by using a neural network on its gradient field.
Later, Rubanova et. al.~\cite{RC19}, proposed a latent ODE model using an ODE-RNN
model as the encoder. The encoder uses neural ordinary differential equations to model the dynamics in the hidden state, and an RNN to update it when a new observation is presented to it. De Brouer et. al.~\cite{DS19}, proposed a continuous time version of the GRU. In~\cite{KM20}, authors propose a neural CDE model which is a continuous analog of an RNN, while Neural ODEs are of ResNET.

Other than the ODE models, there are interpolation models where the entire series is utilized (past and future observations). In~\cite{YZ18,YZ18b}, authors propose a multi-directional RNN for the interpolation that considers near past and near-future observations at a given time point. Sukla et. al.~\cite{SM19}, proposed the Interpolation Network Model, where multiple semi-parametric RBF functions are used for interpolation against a set of reference time points. In~\cite{SM21}, an attention mechanism of the interpolation using the entire data, where observed time points are used as keys and the reference times as queries is proposed. A bidirectional RNN followed by a Variational Autoencoder is used for the classification of the series.

In \cite{pmlr-v119-li20k} the observations are represented as index-value pairs sampled from a continuous but unobserved function to address the missing values. They propose an encoder-decoder framework to learn these sequences. In \cite{10.1145/3459637.3482079} Wang et. al. propose a time-aware Dual-Attention and Memory-Augmented Network leveraging a vector representation of the irregularly sampled data.

In~\cite{HM20}, Horn et. al., proposed Set Functions for Time Series where they model time series as a set of observations, and use a deep set model for the classification. The latent embeddings of each observation are computed using dense layers, and an attention-based aggregation function that has polynomial computational complexity is utilized for aggregating those embeddings. Our proposed model also uses deep set representation of time series. Instead of considering each observation as a set element, we represent each channel as a set element. We use a ResNET~\cite{HZ16} based model for extracting the latent embedding from the given set.
\section{Conclusions}
\label{sec:conc}

In this work, we propose a novel yet simple model, Deep Convolutional Set Function (DCSF), for the classification of asynchronous time series. Specifically, we modify the triplets' set representation of time series into channels' set representation and use a deep sets prediction model for the classification. Moreover, we apply convolutional neural networks, which are state-of-the-art models for the classification of fully observed and equally sampled time series, as the encoder model to extract the embedding of the set elements. Our approach, yields state-of-the-art results on $4$ real world and $2$ synthetic datasets for the tasks of both time series classification and segmentation (online classification) tasks, while providing the better run times compared to a range of state-of-the-art models. The accuracy gains upto $2\%$ for medical datasets, and $30\%$ for synthetic datasets shows the superiority of the proposed model.
\bibliography{bib}
\bibliographystyle{ieeetr}

 \newpage

 \appendix

\section{Hyperparameters searched for the baseline models}

For all the baseline models, we independently tune their hyperparameters on the validation set. 

\spara{GRUDModel:}

\begin{itemize}
    \item Number of Units: \{32, 64, 128, 256, 512, 1024\}
    \item Dropout: \{0.0, 0.1, 0.2, 0.3, 0.4\}
    \item Recurrent Dropout: \{0.0, 0.1, 0.2, 0.3, 0.4\}
\end{itemize}

\spara{Interpolation Networks:}

\begin{itemize}
    \item Units: \{32, 64, 128, 256, 512, 1024\}
    \item Dropout: \{0.0, 0.1, 0.2, 0.3, 0.4\}
    \item Recurrent Dropout: \{0.0, 0.1, 0.2, 0.3, 0.4\}
    \item Imputation Step Size: \{0.5, 1., 2.5, 5.\}
    \item reconstruction fraction: \{0.05, 0.1, 0.2, 0.5, 0.75\}
\end{itemize}

\spara{Phased LSTM}

\begin{itemize}
    \item Number of units: \{32, 64, 128, 256, 512, 1024\}
    \item Use peep holes: \{True, False\}
    \item Leak: \{0.001, 0.005, 0.01\}
    \item Max. wavelen. for initializing hidden nodes: \{10, 100, 1000\}
\end{itemize}

\spara{Latent-ODE}

    \begin{itemize}
        \item rec. dimensions: \{10,30,50\}
        \item rec. layers:   \{1,2,3,4,5\}
        \item gen. layers:  \{1,2,3,4,5\}
        \item Units:  \{25,50,100,150\}
    \end{itemize}

\spara{Seft-ATTN}
    \begin{itemize}
        \item phi layers: \{1,2,3,4,5\}
        \item phi width: \{16, 32, 64, 128, 256, 512\}
        \item phi dropout: \{0.0, 0.1, 0.2, 0.3\}
        \item Att. dropout: \{0.0, 0.1, 0.25, 0.5\}
        \item latent width: \{32, 64, 128, 256, 512, 1024, 2048\}
        \item rho layers: \{1, 2, 3, 4, 5\}
        \item rho width: \{16, 32, 64, 128, 256, 512\}
        \item rho dropout: \{0.0, 0.1, 0.2, 0.3\}
        \item max timescale: \{10, 100, 1000\}
        \item positional dims: \{4, 8, 16\}
    \end{itemize}

\spara{mTAN}
\begin{itemize}
    \item rec. layers: \{32, 64, 128, 256\}
    \item gen. layers: \{25, 50, 100, 150\}
    \item latent dimensions: \{20, 30, 40, 50\}
    \item reference points: \{8,16,32,64,128\}
    \item attention heads: \{1, 2, 4\}
\end{itemize}

\spara{DCSF}
For DCSF, we use the residual block similar to that of ResNET model for time series. For the first residual block, we use 64 filters, and for the subsequent ones, we use 128. In each residual block, we have $3$ CNN layers, and the kernel lengths are 8, 5, and 3 respectively.
\begin{itemize}
    \item Residual blocks: \{1,2,3,4\}
    \item dense layers: \{0, 1, 2, 3, 4, 5\}
    \item dense width: \{32, 64, 128, 256, 512\}
    \item dense dropout: \{0.0, 0.1, 0.2, 0.3\}
\end{itemize}

\spara{RESNET}
\begin{itemize}
    \item Residual blocks: \{3\}
\end{itemize}

\spara{RESNET-forward}
\begin{itemize}
    \item Residual blocks: \{3\}
\end{itemize}

\section{Implementation Details}

Other than mTAN, and Latent ODE, we run all the baseline models using tensorflow 2.6, and trained on NVIDIA GeForce RTX 3090 GPUs. mTAN and Latent ODE are developed on pytorch, and we use torch for implementing them. 

Because of the high complexity and order of magnitude more run time, we could not search for the hyperparamenters of Latent ODE model, instead we use the results from the published papers. It could be possible that the results of Latent ODE are undermined for the medical datasets.

\section{Source code}

Horn et. al. provided a pipeline for training the baseline models\footnote{\url{https://github.com/BorgwardtLab/Set_Functions_for_Time_Series}}. We developed our code in line with that pipeline. Source code with the best hyperparameters for the proposed model can be find at \url{https://github.com/yalavarthivk/DCSF}.

\end{document}